\def\aaaianonymous{false}
\documentclass[letterpaper]{article} % DO NOT CHANGE THIS

\usepackage{aaai2026}
\usepackage{times}  % DO NOT CHANGE THIS
\usepackage{helvet}  % DO NOT CHANGE THIS
\usepackage{courier}  % DO NOT CHANGE THIS
\usepackage[hyphens]{url}  % DO NOT CHANGE THIS
\usepackage{graphicx} % DO NOT CHANGE THIS
\def\UrlFont{\rm}  % DO NOT CHANGE THIS
\usepackage{natbib}  % DO NOT CHANGE THIS AND DO NOT ADD ANY OPTIONS TO IT
\usepackage{caption} % DO NOT CHANGE THIS AND DO NOT ADD ANY OPTIONS TO IT
\usepackage{algorithm}
\usepackage{algorithmic}
\usepackage{amsmath}
\usepackage{amsfonts}
\usepackage{mathrsfs}
\usepackage{booktabs}
\usepackage{adjustbox}
\usepackage{graphicx} 
\usepackage{colortbl}
\usepackage{xcolor}
\usepackage{tabularx}
\usepackage{multirow} 
\usepackage{xcolor}
\usepackage{pifont}
\usepackage{newfloat}
\usepackage{listings}
\DeclareCaptionStyle{ruled}{labelfont=normalfont,labelsep=colon,strut=off} % DO NOT CHANGE THIS
\floatstyle{ruled}
\newfloat{listing}{tb}{lst}{}
\floatname{listing}{Listing}

\ifdefined\aaaianonymous
    \title{AAAI Press Anonymous Submission\\Instructions for Authors Using \LaTeX{}}
\else
    \title{AAAI Press Formatting Instructions \\for Authors Using \LaTeX{} --- A Guide}
\fi

\title{Opt3DGS: Optimizing 3D Gaussian Splatting with Adaptive Exploration and Curvature-Aware Exploitation}
\author {
    Ziyang Huang\textsuperscript{\rm 1}, 
    Jiagang Chen\textsuperscript{\rm 1}, 
    Jin Liu\textsuperscript{\rm 2}, 
    Shunping Ji\textsuperscript{\rm 1,\ding{41}}
}
\affiliations {
    \textsuperscript{\rm 1}Wuhan University
    \textsuperscript{\rm 2}Hangzhou Dianzi University\\
    \{huangziyang2024, chenjiagang2015, jishunping\}@whu.edu.cn, jinliu@hdu.edu.cn
}
\usepackage{bibentry}
\begin{document}

\maketitle

\begin{abstract}
3D Gaussian Splatting (3DGS) has emerged as a leading framework for novel view synthesis, yet its core optimization challenges remain underexplored. We identify two key issues in 3DGS optimization: entrapment in suboptimal local optima and insufficient convergence quality. To address these, we propose Opt3DGS, a robust framework that enhances 3DGS through a two-stage optimization process of adaptive exploration and curvature-guided exploitation. In the exploration phase, an Adaptive Weighted Stochastic Gradient Langevin Dynamics (SGLD) method enhances global search to escape local optima. In the exploitation phase, a Local Quasi-Newton Direction-guided Adam optimizer leverages curvature information for precise and efficient convergence. Extensive experiments on diverse benchmark datasets demonstrate that Opt3DGS achieves state-of-the-art rendering quality by refining the 3DGS optimization process without modifying its underlying representation.
\end{abstract}

\section{1 Introduction}

3D Gaussian Splatting~\cite{kerbl20233d} has recently emerged as a dominant method in novel view synthesis, significantly advancing scene modeling through its superior representational capabilities. Unlike implicit predecessors such as Neural Radiance Fields (NeRF)~\cite{mildenhall2021nerf}, 3DGS leverages an explicit approach, modeling the radiance field of scenes using discrete Gaussian primitives. This explicit representation offers considerable advantages in computational efficiency and modeling flexibility, facilitating widespread adoption in diverse applications such as geometric reconstruction~\cite{yu2024gaussian, chen2024pgsr, guedon2024sugar}, simultaneous localization and mapping (SLAM)~\cite{matsuki2024gaussian}, semantic scene understanding~\cite{cen2025segment}, and dynamic scene modeling~\cite{yang2023real}.
\begin{figure}[ht!]
    \centering
    \includegraphics[width=\linewidth]{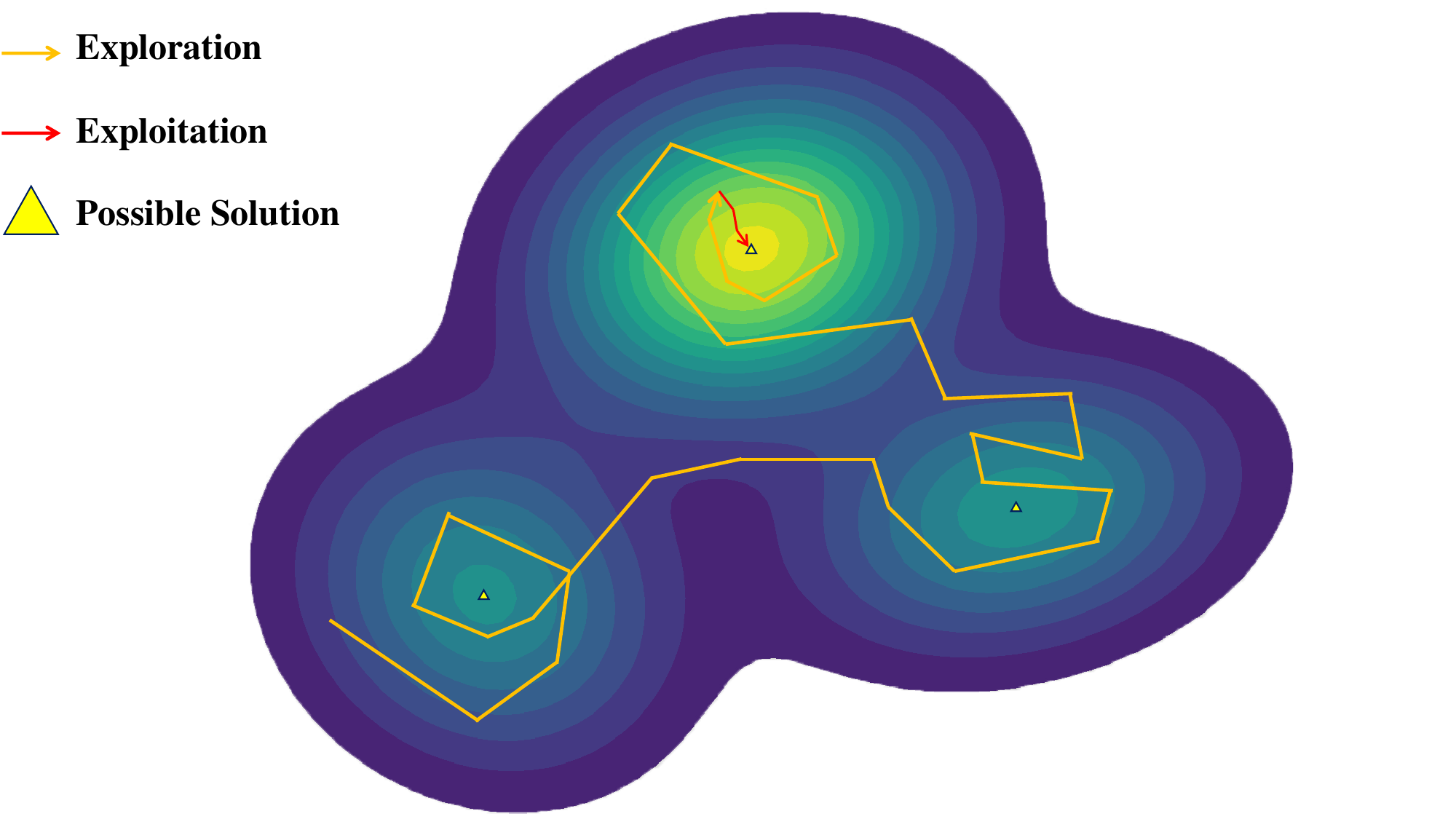}
    \caption{Exploration and Exploitation. The exploration stage promotes global search across modes using Adaptive Weighted SGLD, while the exploitation stage enables precise, curvature-aware convergence with Local Quasi-Newton direction-guided Adam optimizer.}
    \label{fig:exploration-exploitation}
\end{figure}

% Despite the rapid adoption of 3DGS, practitioners often encounter unstable optimization process in complex scenes: Gaussian primitives tend to gather around salient structures (e.g., foreground objects), while background or geometrically intricate regions remain under-reconstructed. This empirical observation suggests that effectively optimizing Gaussian primitives to reconstruct a realistic radiance field remains a challenging non-convex optimization problem.

Despite these advantages, effectively optimizing Gaussian primitives to reconstruct a radiance field remains a challenging non-convex optimization problem. Non-convex optimization inherently faces the risk of convergence to local optima, complicating the attainment of globally optimal scene representations. The original 3DGS method employs heuristic rules within its adaptive density control (ADC) step, using fixed thresholds to guide Gaussian cloning, splitting, and pruning. Yet, these heuristics lack robustness and frequently yield suboptimal outcomes. Recent advancements, such as 3DGSMCMC~\cite{kheradmand20243d}, attempt to address these limitations by modeling 3DGS optimization as a stochastic gradient Langevin dynamics (SGLD) process, incorporating opacity-based probabilistic sampling for Gaussian addition and removal. Although 3DGSMCMC unifies the training procedure, it does not completely resolve the local optima issue. Specifically, 3DGSMCMC introduces an inherent clustering effect, causing excessive Gaussian accumulation in already well-reconstructed regions. Based on Bayesian theorem, such clustering creates sampling bias, prematurely trapping the optimization in local modes and restricting exploration of the global solution space. Furthermore, standard first-order optimizers, like Adam~\cite{kingma2014adam}, commonly used in existing 3DGS methods lack curvature information, hindering precise convergence during later training stages and limiting reconstruction quality. These limitations motivate the pursuit of optimization frameworks with enhanced exploration and precise convergence to tackle the complex non-convex landscape of 3DGS training.

In this paper, we propose Opt3DGS, a general and effective optimization framework for 3DGS. As shown in Figure 1, our approach divides the training process into two stages: Exploration and Exploitation. In the Exploration stage, inspired by the flat-histogram principle~\cite{wang2001efficient}, we introduce Adaptive Weighted SGLD (AW-SGLD). AW-SGLD flattens the posterior distribution to reduce energy barriers between modes, enabling the model to escape local traps and explore the solution space more thoroughly. This increases the likelihood of identifying the global optimal mode. In the Exploitation stage, once the model approaches a high-quality solution, we design a Local Quasi-Newton Direction-guided Adam optimizer. This optimizer leverages historical gradient information to estimate curvature-aware update directions, achieving more precise convergence than standard Adam while maintaining its robustness. Unlike full Quasi-Newton methods, it avoids the need for computationally expensive line searches. We evaluate Opt3DGS on public benchmarks, including MipNeRF360, Tanks \& Temples, and DeepBlending, and compare it with state-of-the-art 3DGS methods. Experimental results demonstrate that Opt3DGS achieves superior rendering quality, validating the effectiveness of our optimization-centric approach.

Our contributions are summarized as follows:
\begin{itemize}
    \item We analyze the local mode trapping issue in 3DGS from a Bayesian perspective and introduce AW-SGLD to enhance global exploration and improve convergence to the optimal mode.
    \item We overcome the limitation of first-order optimizers in the exploitation phase by developing a Local Quasi-Newton Direction-guided Adam optimizer, enabling precise convergence and enhanced rendering quality.
    \item By focusing solely on optimization without altering the Gaussian representation, Opt3DGS achieves state-of-the-art rendering performance.
\end{itemize}

\section{2 Related Work}
Neural rendering has recently made remarkable progress in novel view synthesis by learning scene representations directly from images. Among these methods, implicit approaches such as NeRF~\cite{mildenhall2021nerf} and explicit approaches like 3DGS~\cite{kerbl20233d} represent two dominant paradigms.

NeRF~\cite{mildenhall2021nerf} pioneered implicit 3D scene representation by learning a volumetric radiance field with an MLP and using volumetric rendering to synthesize novel views. Subsequent extensions improve different aspects of NeRF: NeRF++~\cite{zhang2020nerf++} models unbounded scenes, Mip-NeRF~\cite{barron2021mip} and Mip-NeRF360~\cite{barron2022mip} address aliasing with conical frustum sampling, D-NeRF~\cite{pumarola2021d} incorporates temporal dynamics, and InstantNGP~\cite{muller2022instant} accelerates training with multi-resolution hash encodings. Despite these advances, implicit methods remain computationally expensive due to the need for dense neural evaluations along rays.

3DGS~\cite{kerbl20233d} represents a scene using explicit 3D Gaussian primitives and renders them via parallel rasterization, thereby avoiding the expensive neural field evaluations required by NeRF-based volume rendering~\cite{mildenhall2021nerf}. 
This design achieves real-time novel view synthesis with competitive visual quality. Building on this framework, subsequent works have sought to overcome its limitations from different perspectives.
To improve rendering fidelity, methods such as Mip-Splatting~\cite{yu2024mip} and multi-scale splatting~\cite{yan2024multi} address aliasing.
To enhance expressiveness of representation, 2DGS~\cite{huang20242d} uses 2D surface-aligned Gaussians for higher geometric fidelity. 3DHGS~\cite{li20253d} refine the primitive model with half-Gaussians to better handle discontinuities. SSS~\cite{zhu20253d} introduce Student's T-distribution to improve expressiveness. BBSplat~\cite{svitov2024billboard} using optimizable textured planar primitives to learn RGB textures and alpha maps achieving accurate representation.
Other works focus on efficiency, including anchor-based model compression~\cite{lu2024scaffold}, faster training via resource allocation or Newtonian optimization~\cite{chen2025dashgaussian, lan20253dgs, pehlivan2025second}, sort-free rendering for lightweight devices~\cite{hou2024sort}.
Densification strategies have also been actively explored: AbsGS~\cite{ye2024absgs} and RevisingGS~\cite{rota2024revising} design more principled or error-driven adaptive density control, while  FreGS~\cite{zhang2024fregs} mitigates over-reconstruction caused by densification through frequency-domain regularization. Methods based on Stochastic Gradient Markov Chain Monte Carlo (SGMCMC) unify the update, addition, and removal of Gaussian primitives within a single optimization framework. For instance, 3DGSMCMC~\cite{kheradmand20243d} employ stochastic gradient Langevin dynamics and SSS~\cite{zhu20253d} adopts stochastic gradient Hamiltonian Monte Carlo for enhanced exploration. 

Despite these efforts, stochastic methods still suffer from Gaussian over-clustering and lack mechanisms for precise, curvature-aware convergence, which motivates the optimization framework proposed in this work.

\section{3 Background and Limitation}
This section introduces the basics of 3DGS~\cite{kerbl20233d} and its extension 3DGSMCMC~\cite{kheradmand20243d}, and discusses the key limitations of these methods that motivate our proposed approach.

\subsection{3.1 Preliminary}
3DGS models a scene as a collection of explicit Gaussian primitives.
Each primitive is parameterized by a 3D position $\boldsymbol{\mu} \in \mathbb{R}^3$, 
an opacity scalar $o \in \mathbb{R}$,
view-dependent spherical harmonic coefficients for appearance modeling,
and a covariance matrix $\boldsymbol{\Sigma} = \mathbf{R}\mathbf{s}\mathbf{s}^{\top}\mathbf{R}^{\top}$ that determines the spatial extent and orientation of the Gaussian,
where $\mathbf{s} \in \mathbb{R}^3$ is a scale vector for the axis lengths and 
$\mathbf{r} \in \mathbb{R}^4$ (represented as a quaternion) for the rotation matrix $\mathbf{R}$.
During rendering, each 3D Gaussian is projected onto the image plane, and the final pixel color is computed using alpha blending:
\begin{equation}
\mathbf{c}(\mathbf{x}) = \sum_{i=1}^N \mathbf{c}_i \cdot o_i \cdot T_i,
\quad  T_{i+1} = (1 - o_i) \cdot T_i,
\label{eq:final_color}
\end{equation}
where $\mathbf{c}_i$, $o_i$, and $T_i$ denote the color, opacity, and accumulated transmittance of the $i$-th Gaussian, respectively.

The adaptive density control in vanilla 3DGS, while effective, lacks robustness in certain scenarios. To address this limitation, the 3DGSMCMC framework reformulates 3DGS as a Markov Chain Monte Carlo (MCMC) process, treating each Gaussian primitive as a sample drawn from the scene’s posterior distribution. Instead of stochastic gradient descent (SGD), 3DGSMCMC employs Stochastic Gradient Langevin Dynamics (SGLD) for parameter updates:
\begin{equation}
\begin{aligned}
g_k &\leftarrow g_{k-1}
- \lambda_{lr} \cdot \nabla_g \mathbb{E}_{I \sim \mathcal{I}}
\big[ \mathcal{L}_{\text{total}}(g_{k-1}; I) \big]
+ \lambda_{\text{noise}} \cdot \epsilon, \\
\epsilon &= \lambda_{lr} \cdot \sigma(-k(t-o)) \cdot \Sigma \eta,
\qquad \epsilon = [\epsilon_\mu, 0],
\end{aligned}
\label{eq:sgld_update}
\end{equation}
where $g_k$ denotes the Gaussian parameters at iteration $k$, 
$\lambda_{lr}$ the learning rate, 
$\mathcal{L}_{\text{total}}$ the total loss, 
and $\lambda_{\text{noise}}$ the coefficient controlling the injected noise. 
The term $\epsilon$ is parameterized by the sigmoid function $\sigma(\cdot)$ with hyperparameters $k$ and $t$, 
the Gaussian opacity $o$, the covariance matrix $\Sigma$, 
and a 3D standard Gaussian random vector $\eta$.

When adding Gaussians, 3DGSMCMC samples the locations of the new Gaussians from the normalized opacity-based probability distribution of the current Gaussian set. For pruning Gaussians, 3DGSMCMC reloates the discarded Gaussians to the location of high-opacity Gaussians through the same opacity-based sampling.
To maintain the stability of the Markov chain, the parameters of the newly generated Gaussians are computed as:
\begin{equation}
\resizebox{\linewidth}{!}{$
\begin{aligned}
\boldsymbol{\Sigma}_{1,\dots,N}^{\text{new}} &= \left( o^{\text{old}} \right)^2 
\left( \sum_{i=1}^{N} \sum_{k=0}^{i-1} 
\left( \binom{i-1}{k} \frac{(-1)^k (o^{\text{new}})^{k+1}}{\sqrt{k+1}} \right) 
\right)^{-2} \boldsymbol{\Sigma}^{\text{old}}, \\
&\mu_{1,\dots,N}^{\text{new}} = \mu^{\text{old}}, \quad 
o_{1,\dots,N}^{\text{new}} = 1 - \sqrt[N]{1 - o^{\text{old}}}
\end{aligned}
$}
\label{eq:new_covariance}
\end{equation}

To promote sparsity in opacity and control the scale of the covariance matrices, 3DGSMCMC introduces two additional regularization term in loss function:
\begin{equation}
\begin{aligned}
L_{total} =(1-\lambda_{ssim})\times L_1 + \lambda_{ssim} \times L_{ssim} + \\\lambda_o\times \sum_i|o_i|_1+\lambda_{\Sigma}\times\sum_{ij}\left|\sqrt{\mathrm{eig}_j(\Sigma_i)}\right|_1
\end{aligned}
\label{eq:3DGSMCMCLoss}
\end{equation}

\subsection{3.2 Limitation of Existing Framework}

3DGSMCMC formulates 3DGS optimization as a Markov Chain Monte Carlo process and achieves promising results, as the Langevin dynamics component encourages exploration of the posterior distribution. 
However, reconstructing complex scenes remain challenging.
The posterior energy landscape is often highly multi-modal, with each mode corresponding to a different Gaussian configuration that explains the scene.
When the energy barriers between modes are large, the combination of gradient guidance and Langevin noise is often insufficient to push the model out of its current local mode.

This issue is further exacerbated by the opacity-based sampling mechanism used in 3DGSMCMC. When adding or relocating Gaussians, a set of new sample positions is drawn independently and identically distributed (i.i.d.) from a probability distribution $\pi(x)$, which is proportional to the normalized opacities of the current Gaussians:
\begin{equation}
x^{(1)},\, x^{(2)},\, \ldots,\, x^{(N)} \overset{\mathrm{i.i.d.}}{\sim} \pi(x)
\end{equation}
Here each $x^{(i)}$ corresponds to the spatial position where a new Gaussian will be placed. Although straightforward, this opacity-driven rule induces a clustering effect: new Gaussians tend to accumulate in regions that were discovered early in training. As dominant structures become highly opaque, subsequent sampling becomes more biased toward these well-reconstructed areas, leaving under-explored or geometrically complex regions insufficiently covered. From an MCMC perspective, this bias limits efficient exploration and traps the chain in a single posterior mode.

\section{4 Method}
To address the prevalent issues of local mode trapping and limited convergence in 3DGS optimization, we propose Opt3DGS, a novel optimization framework that combines two complementary components: 
an Adaptive Weighted SGLD that promotes global exploration and helps the model escape local minima in exploration stage; A Local Quasi-Newton Direction-guided Adam that refines the solution in exploitation stage with more accurate, curvature-aware updates. The following sections describe these two components in detail.

\subsection{4.1 3DGS with Adaptive Weighted SGLD~\label{sec_awmcmc}}
Despite the clustering effect of opacity-based sampling (Sec. 3.2), it still offers favorable initializations for new Gaussians. Instead of modifying the sampling mechanism, we enhance the model’s exploration to avoid premature convergence to suboptimal modes.
A direct way to enhance exploration is to increase the Langevin noise intensity $\lambda_{\text{noise}}$. However, this is not robust: scene complexity varies, and excessive noise can destabilize training and impede convergence.

To address this, we introduce the flat histogram principle and propose an adaptive weighted stochastic gradient langevin dynamics(AW-SGLD) update for 3DGS. Let the configuration of Gaussian primitives for current scene follow a probability distribution $\mathrm{P}(g)$, defined as:
\begin{equation}
\mathrm{P}(g) \propto \exp\left(-\frac{\mathcal{L}_{\text{total}}(g)}{\tau}\right), \quad g \in \mathcal{G},
\label{eq:pi_distribution}
\end{equation}
where $g$ denotes the current sample, $\mathcal{G}$ is the sample space, $\mathcal{L}_{\text{total}}$ is the total training loss (the energy function of the target distribution), and $\tau$ is the temperature parameter.

Our objective is to construct a flattened distribution $\rho(g)$ based on $\mathrm{P}(g)$ to facilitate the traversal of the sample space. To achieve this, we divide $\mathcal{G}$ into $m$ disjoint subregions based on the energy levels of $\mathcal{L}_{\text{total}}(g)$:
\begin{equation}
\resizebox{\linewidth}{!}{$
\mathcal{G} = \mathcal{G}_1 \cup \mathcal{G}_2 \cup \dots \cup \mathcal{G}_m,\quad 
\mathcal{G}_n = \{ g : u_{n-1} < \mathcal{L}_{\text{total}}(g) < u_n \}
$}
\label{eq:partition}
\end{equation}
where $u_0 = -\infty, \ u_m = +\infty$, while $u_1$ and $u_{m-1}$ are specified by the user. Inspired by~\cite{neal2001annealed}, we define the flattened distribution $\rho(g)$ as:
\begin{equation}
\rho(g) \propto \frac{\mathrm{P}(g)}{\Psi^{\zeta}(\Theta, \mathcal{L}_{\text{total}}(g))},
\label{eq:flattened_distribution}
\end{equation}
where $\zeta > 0$ is a flattening hyperparameter controlling the degree of flattening, and $\Psi(\Theta, \mathcal{L}_{\text{total}}(g))$ is a weighting function that takes the energy of current sample $\mathcal{L}_{\text{total}}(g)$, and a weight vector $\Theta$ as inputs, and returns the corresponding flattening weight, where $\Theta$ is defined as:
\begin{equation}
\begin{split}
\Theta =& \bigl\{ (\theta(1), \theta(2), \dots, \theta(m)) \mid \\
&0 < \theta(1), \theta(2), \dots, \theta(m) < 1, \sum_{i=1}^m \theta(i) = 1 \bigr\},
\end{split}
\label{eq:weight_vector}
\end{equation}
with $\theta(i) = 1/m$ at the start of training.

To avoid gradient vanishing issues associated with the piecewise constant form of $\Psi$, as met in~\cite{liang2007stochastic}, we construct $\Psi$ using a piecewise exponential interpolation function, following~\cite{deng2020contour}:
\begin{equation}
\resizebox{\linewidth}{!}{$
\begin{split}
\Psi(\Theta, \mathcal{L}_{\text{total}}(g)) =&\sum_{i=1}^m \mathbf{1}_{\{u_{i-1} \leq \mathcal{L}_{\text{total}}(g) \leq u_i\}} \times\\ 
& \theta(i-1) 
\exp\left( (\log \theta(i) - \log \theta(i-1)) \frac{\mathcal{L}_{\text{total}}(g) - u_{i-1}}{\Delta u} \right)
\end{split}
$}
\label{eq:weighting_function}
\end{equation}
where $1_A$ is the indicator function, equal to 1 when event $A$ occurs and 0 otherwise. This formulation interpolates the discrete $\theta(i)$ values exponentially based on the energy of the current sample $g$.

To derive the update rule for the flattened distribution $\rho(g)$, we compute the gradient:
\begin{equation}
\resizebox{\linewidth}{!}{$
\nabla_g \log \rho(g) = -\frac{\nabla_g \mathcal{L}_{\text{total}}(g)}{\tau} \times \left[ 1 + \zeta \tau \frac{\partial \log \Psi(\Theta, \mathcal{L}_{\text{total}}(g))}{\partial \mathcal{L}_{\text{total}}(g)} \right] .
$}
\label{eq:gradient_varpi}
\end{equation}
The format \eqref{eq:gradient_varpi} is expanded to:
\begin{equation}
\begin{split}
\nabla_g &\log \rho(g) = -\frac{\nabla_g \mathcal{L}_{\text{total}}(g)}{\tau} \times  \\ &\biggl[ 1 + \zeta \tau \frac{\log \theta(J(g)) - \log (\theta(J(g) - 1) \vee 1)}{\Delta u} \biggr]
\end{split}
\label{eq:simplified_gradient}
\end{equation}
where $\Delta u = u_n - u_{n-1}$ for $n \in \{2, \dots, m-1\}$ and $J(g) \in \{1, 2, \dots, m\}$ denotes the index of the subregion containing the current sample $g$:
\begin{equation}
J(g) = \sum_{i=1}^m i \, 1_{u_{i-1} < \mathcal{L}_{\text{total}}(g) \leq u_i}.
\label{eq:subregion_index}
\end{equation}
Compared to the gradient under the original distribution, ~\eqref{eq:simplified_gradient} introduces an additional gradient multiplier $\nu$:
\begin{equation}
\nu = 1 + \zeta \tau \frac{\log \theta(J(g)) - \log (\theta(J(g) - 1) \vee 1)}{\Delta u}.
\label{eq:gradient_multiplier}
\end{equation}

To align the update with the flattened distribution $\rho(g)$, the gradient multiplier is merged into the SGLD update (2):
\begin{equation}
\resizebox{\linewidth}{!}{$
\begin{split}
g_k \leftarrow g_{k-1} &- \lambda_{\text{lr}}\cdot \nu \cdot \nabla_g \mathbb{E}_{I \sim \mathcal{I}} \left[ \mathcal{L}_{\text{total}}(g_{k-1}; I) \right] + \lambda_{\text{noise}} \cdot \epsilon,
\end{split}
$}
\label{eq:flattened_update}
\end{equation}

\begin{figure}[ht!]
    \centering
    \includegraphics[width=\linewidth]{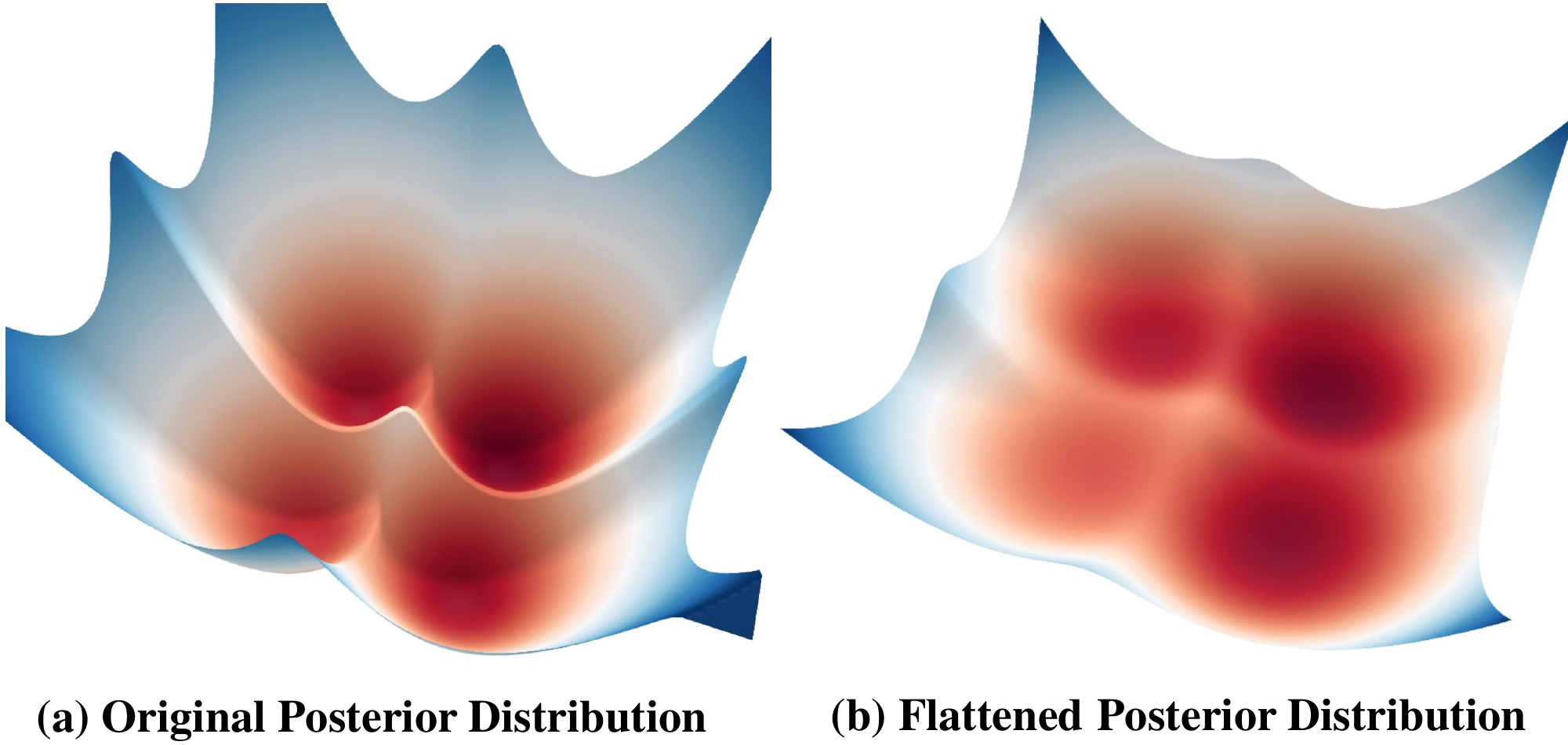}
    \caption{Original (a) and Flattened (b) posterior distribution. In the original distribution, High energy barriers between modes can trap the model in a single basin. The flattened distribution reduces these barriers, enabling free exploration across modes.}
    \label{fig:exploration-exploitation}
\end{figure}

Next, we update the weight vector $\Theta$ to ensure that $\Psi(\Theta, \mathcal{L}_{\text{total}}(g))$ produces appropriate flattening weights as training goes. Following~\cite{deng2020contour}, we employ a stochastic approximation (SA) approach to estimate $\Theta$ during training. The goal of SA is to drive each $\theta(i)$ to converge to the cumulative probability density of the corresponding subregion under the original distribution. At each iteration, before updating the Gaussian primitives, we perform:
\begin{equation}
\theta_k(i) = \theta_{k-1}(i) + \lambda_{\theta} \theta_{k-1}^{\zeta}(J(g_k)) \cdot \left( 1_{i = J(g_k)} - \theta_{k-1}(i) \right),
\label{eq:theta_update}
\end{equation}
where $\lambda_{\theta}$ is the learning rate for updating $\theta(i)$. This step increases the weight of the current subregion $J(g_k)$ while decreasing the weights of other subregions.

During exploration stage, we simultaneously update the Gaussian primitives $g$ and the weight vector $\Theta$ using the above rules. Consistent with 3DGSMCMC, we employ the gradients from Adam optimizer in place of raw gradients in \eqref{eq:flattened_update} to enhance optimization stability. By flattening the posterior distribution, AW-SGLD enhances the exploration capability of the model, as illustrated in Figure 2, thereby increasing the likelihood of converging to the deepest mode—the one containing the optimal solution.

\begin{table*}[ht!]
\centering
\adjustbox{width=\linewidth}{
\begin{tabular}{lccccccccc}
\toprule
\multirow{2}{*}{\textbf{Methods}} & \multicolumn{3}{c}{\textbf{MipNeRF360}} & \multicolumn{3}{c}{\textbf{Tanks \& Temples}} & \multicolumn{3}{c}{\textbf{DeepBlending}} \\
\cmidrule(lr){2-4} \cmidrule(lr){5-7} \cmidrule(lr){8-10}
\multicolumn{1}{c}{} & \textbf{PSNR($\uparrow$)} & \textbf{SSIM($\uparrow$)} & \textbf{LPIPS($\downarrow$)} & \textbf{PSNR($\uparrow$)} & \textbf{SSIM($\uparrow$)} & \textbf{LPIPS($\downarrow$)} & \textbf{PSNR($\uparrow$)} & \textbf{SSIM($\uparrow$)} & \textbf{LPIPS($\downarrow$)} \\
\midrule
MipNeRF & 29.23 & 0.844 & 0.207 & 22.22 & 0.759 & 0.257 & 29.40 & 0.901 & 0.245 \\
3DGS & 28.69 & 0.870 & 0.182 & 23.14 & 0.841 & 0.183 & 29.41 & 0.903 & 0.243 \\
Scaffold-GS & 28.84 & 0.848 & 0.220 & 23.96 & 0.853 & 0.177 & \textbf{30.21} & 0.906 & 0.254 \\
FreGS & 27.85 & 0.826 & 0.209 & 23.96 & 0.841 & 0.183 & 29.93 & 0.904 & \underline{0.240} \\
3DHSGS & 29.56 & 0.873 & 0.178 & 24.49 & 0.857 & 0.169 & 29.76 & 0.905 & 0.242 \\
3DGSMCMC & 29.89 & \textbf{0.900} & 0.190 & 24.29 & 0.860 & 0.190 & 29.67 & 0.900 & 0.320 \\
SSS & \underline{29.90} & 0.893 & \underline{0.145} & \textbf{24.87} & \underline{0.873} & \textbf{0.138} & 30.07 & \underline{0.907} & 0.247 \\
Ours & \textbf{29.96} & \underline{0.897} & \textbf{0.143} & \underline{24.80} & \textbf{0.875} & \underline{0.139} & \underline{30.09} & \textbf{0.911} & \textbf{0.229} \\
\bottomrule
\end{tabular}
}
\caption{Quantitative comparison between ours and baseline methods. For a fair comparison, we use the same resolution settings and maximum number of Gaussians as in 3DGSMCMC. The \textbf{best} and \underline{second-best} results in each column are highlighted.}
\label{tab:metrics}
\end{table*}

\subsection{4.2 Local Quasi-Newton Direction-guided Adam optimizer ~\label{sec_awadam}}
Although the flattened distribution based on the flat-histogram principle enhances the model’s exploration capability and helps escape from local traps, the ultimate goal of 3DGS remains to find the global optimum that minimizes the loss. Enhanced exploration alone does not guarantee precise convergence to the optimal point within a mode.

To improve convergence quality in the later stages of training, we switch to exploitation from exploration, and design a curvature-aware optimization strategy. While prior works have employed Newton’s method~\cite{lan20253dgs} or the Levenberg–Marquardt (LM)~\cite{hollein20243dgs} algorithm to accelerate optimization, these approaches require complicated calculations of the Hessian matrix or its approximation. Our objective is achieving precise convergence without incurring excessive computational overhead.

We propose a Local Quasi-Newton Direction-guided Adam Optimizer (LQNAdam). Here, ``local'' indicates that each Gaussian primitive is treated independently. We apply the limited-memory Broyden-Fletcher-Goldfarb-Shanno (L-BFGS)~\cite{nocedal2006numerical} algorithm to the positional attributes $\mu$ of each Gaussian primitive and estimate a quasi-Newton direction based on the past $K$ steps. Following \cite{liu1989limited}, the value of $K$ is chosen between $3$ and $7$ to balance computational cost and performance. This is motivated by the observation in 3DGS\textsuperscript{2}~\cite{lan20253dgs} that positional attributes significantly influence rendering quality and that Gaussians are weakly coupled, enabling parallel estimation of local quasi-Newton directions. Details of the L-BFGS algorithm are provided in the Supplementary Material.

To ensure stable convergence, L-BFGS typically uses a line search to determine the step size. However, performing line search for each Gaussian primitive is impractical in our context. Instead, we treat the Quasi-Newton direction from L-BFGS as a pseudo-gradient and feed it into the Adam optimizer, computing the final update direction. This approach leverages Adam’s robustness while incorporating curvature-aware directions, yielding more accurate updates as the model approaches a solution. Notably, L-BFGS does not require computation of the Hessian matrix, making our optimizer compatible with various loss functions.

Let $\mathbb{D}$ denote the quasi-Newton direction estimated by L-BFGS for a Gaussian’s positional attributes $\mu$. The final update direction is computed as $\textit{Adam}(\mathbb{D})$. Within the Markov Chain Monte Carlo framework, the update rule for our Local Quasi-Newton Direction-guided Adam Optimizer is:
\begin{equation}
\mu_{t+1} = \mu_t - \lambda_{\text{lr}} \cdot \textit{Adam}(\mathbb{D}) + \lambda_{\text{noise}} \cdot \epsilon_\mu.
\label{eq:quasi_newton_update}
\end{equation}

In the later stages of training, we switch to the exploitation. In this stage, the L1 loss is replaced by L2 loss, and LQNAdam is adopted. We also disable the gradient multiplier $\nu$, allowing model updates to follow the original distribution, thereby focusing on enhancing convergence quality.

 \section{5 Experiments}

\begin{figure*}[ht!]
    \centering
    \includegraphics[width=\linewidth]{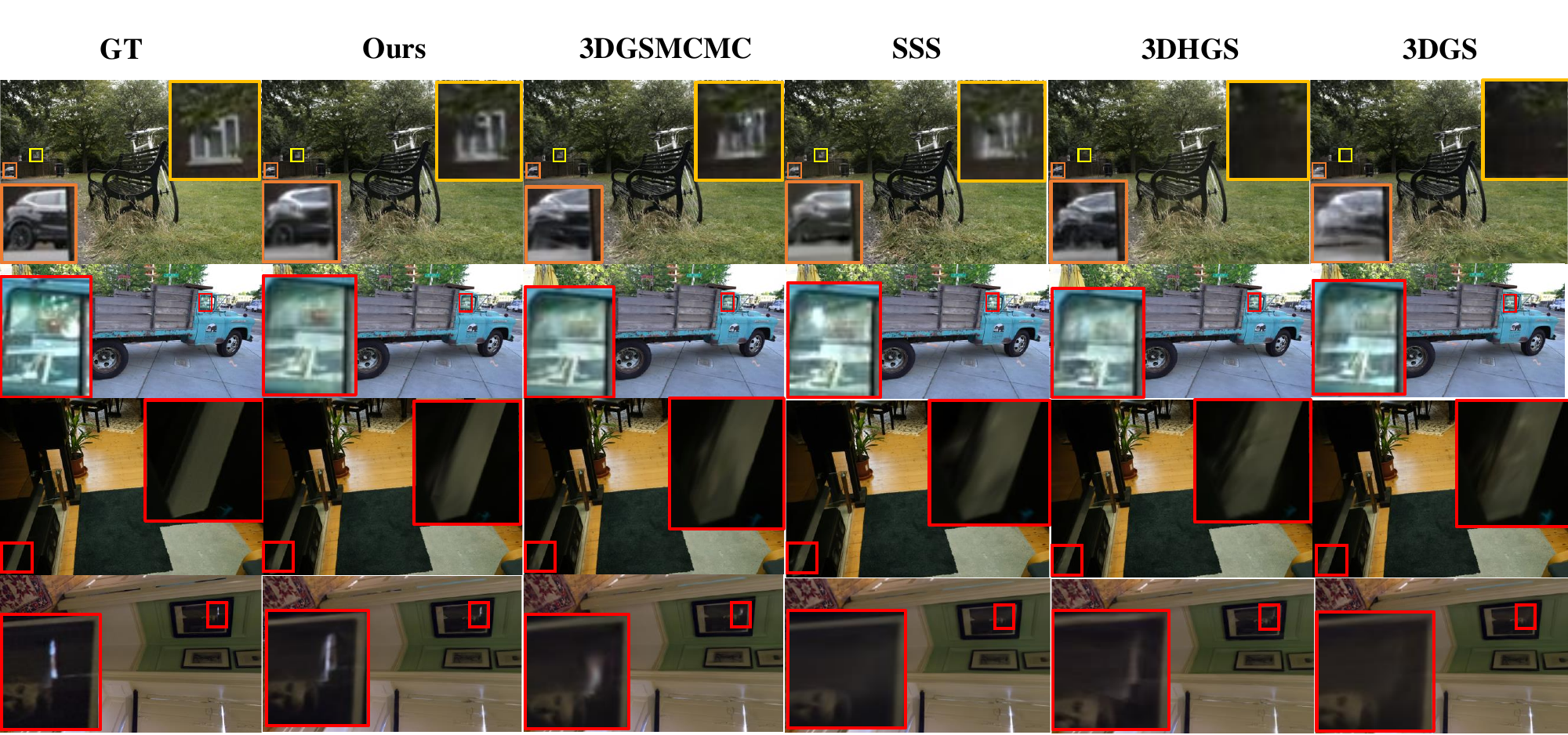}
    \caption{Visualization comparison. Our method achieves higher fidelity in challenging regions like distant and fine details.}
    \label{fig:viz_com}
\end{figure*}

\subsubsection*{Implementation Details}

All experiments with Opt3DGS are performed on a NVIDIA RTX 4090 GPU, with a total of 30{,}000 optimization iterations. 
The growth rate of Gaussian primitives is fixed at 5\%, following the setting in 3DGSMCMC. 
In the Adaptive Weighted SGLD module, the posterior energy range is discretized into 200 uniform bins. 
For all scenes the energy interval is set to $[0.0, 0.2]$, except for the \textit{train} scene in the Tanks \& Temples dataset, where a wider range of $[0.0, 0.3]$ is used. 
A warm-up of 2{,}500 iterations is applied to stabilize energy estimates before adaptive weighting, and the flattening coefficient is fixed to $\zeta = 0.75$ for all experiments. 
For the quasi-Newton updates, we employ an L-BFGS history size $K$ of 5, and compute quasi-Newton directions for Gaussian positions in parallel on CUDA. 
The training process switches from the exploration phase to the exploitation phase at iteration 29{,}000, and the final 1{,}000 iterations are used for exploitation for fine convergence.

\subsubsection*{Baseline Methods}

We compare Opt3DGS with vanilla 3DGS~\cite{kerbl20233d} and several representative variants that aim to improve the optimization or representation of 3DGS: 3DHGS~\cite{li20253d}, FreGS~\cite{zhang2024fregs}, Scaffold-GS~\cite{lu2024scaffold}, 3DGSMCMC~\cite{kheradmand20243d}, and SSS~\cite{zhu20253d}, as well as MipNeRF~\cite{barron2022mip} as a representative NeRF-based method. 
Among these baselines, 3DHGS and SSS improve the expressiveness of Gaussian primitives; Scaffold-GS introduces an MLP component to accelerate training and enhance quality; 3DGSMCMC employs a stochastic optimization framework based on MCMC; and FreGS applies frequency regularization to boost both convergence speed and rendering fidelity.
All reported baseline results are taken from their original publications.

\subsubsection*{Datasets and Metrics}

Following prior work on 3DGS, we evaluate the proposed Opt3DGS on three widely used real-world datasets: 
\textbf{MipNeRF360}~\cite{barron2022mip}, which contains 3 outdoor scenes (\textit{garden}, \textit{bicycle}, \textit{stump}) and 4 indoor scenes (\textit{kitchen}, \textit{bonsai}, \textit{room}, \textit{counter}); 
\textbf{DeepBlending}~\cite{hedman2018deep}, consisting of 2 indoor scenes (\textit{drjohnson} and \textit{playroom}); 
and \textbf{Tanks \& Temples}~\cite{knapitsch2017tanks}, with 2 outdoor scenes (\textit{train} and \textit{truck}). 
For quantitative evaluation, we adopt three widely used visual quality metrics~\cite{zhang2018unreasonable}: PSNR, SSIM, and LPIPS, computed on the test images.

\begin{table*}[ht!]
\centering
\adjustbox{width=\linewidth}{
\begin{tabular}{lccccccccc}
\toprule
\multirow{2}{*}{\textbf{Methods}}  & \multicolumn{3}{c}{\textbf{MipNeRF360}} & \multicolumn{3}{c}{\textbf{Tanks \& Temples}} & \multicolumn{3}{c}{\textbf{DeepBlending}} \\
\cmidrule(lr){2-4} \cmidrule(lr){5-7} \cmidrule(lr){8-10}
& \textbf{PSNR($\uparrow$)} & \textbf{SSIM($\uparrow$)} & \textbf{LPIPS($\downarrow$)} & \textbf{PSNR($\uparrow$)} & \textbf{SSIM($\uparrow$)} & \textbf{LPIPS($\downarrow$)} & \textbf{PSNR($\uparrow$)} & \textbf{SSIM($\uparrow$)} & \textbf{LPIPS($\downarrow$)} \\
\toprule
3DGS & 27.89 & 0.840 & 0.260 & 21.93 & 0.800 & 0.270 & 29.55 & 0.900 & 0.330 \\
3DGSMCMC & 29.72 & 0.890 & 0.190 & 24.21 & 0.860 & 0.190 & 29.71 & 0.900 & 0.320 \\
Ours &  \textbf{29.78} & \textbf{0.893} & \textbf{0.149} & \textbf{24.39} & \textbf{0.865} & \textbf{0.151} & \textbf{29.90} & \textbf{0.905} & \textbf{0.236} \\
\bottomrule
\end{tabular}
}
\caption{Quantitative comparison between our method and baselines with random initialization. Although random initialization leads to a poor starting state and makes optimization challenging, our method achieves superior results across all metrics.}
\label{tab:metrics}
\end{table*}

\subsection{5.1 Benchmark Results}

The quantitative comparison with various baselines across three benchmark datasets is summarized in Table 1. Our Opt3DGS achieves the best performance on 5 out of 9 metrics and ranks second on the remaining 4. 
Compared to 3DGSMCMC, which shares the same Gaussian representation but differs solely in the optimization strategy, Opt3DGS achieves consistent performance gains across all metrics except for the SSIM on the MipNeRF360 dataset. On the Tanks \& Temples dataset, we achieve PSNR/SSIM/LPIPS scores of 24.80 / 0.875 / 0.139, compared to 24.29 / 0.860 / 0.190 for 3DGSMCMC, representing improvements of 2.09$\%$, 1.74$\%$, and 26.84$\%$ respectively. Compared to the previous state-of-the-art SSS, which improves both the 3DGS representation and the training optimization, we achieve comparable or better results. On the DeepBlending dataset, we achieve 30.09 / 0.911 / 0.229, compared to 30.07 / 0.907 / 0.247 for SSS, representing improvements of 0.06$\%$, 0.4$\%$, and 7.2$\%$ respectively. These results confirm that better posterior exploration and convergence, rather than architectural modifications, can yield substantive performance improvements.
We present qualitative comparisons of the novel view synthesis in Figure 3. We compare our method with several baselines: 3DGSMCMC, SSS, 3DHGS, and 3DGS. For a fair comparison, our method adopts the same configuration as 3DGSMCMC, and the other methods use their default settings. We show results on four novel views, where our method demonstrates superior rendering fidelity, particularly in distant background regions, fine geometric details, and subtle scene structures that are difficult to capture.

\begin{table*}[ht!]
\centering
\adjustbox{width=\linewidth}{
\begin{tabular}{lcccccccc}
\toprule
\multirow{2}{*}{\textbf{Methods}} & \multicolumn{4}{c}{\textbf{Train}} & \multicolumn{4}{c}{\textbf{Truck}} \\
\cmidrule(lr){2-5} \cmidrule(lr){6-9}
\multicolumn{1}{c}{\textbf{}} & \textbf{PSNR($\uparrow$)} & \textbf{SSIM($\uparrow$)} & \textbf{LPIPS($\downarrow$)} & \textbf{Time} & \textbf{PSNR($\uparrow$)} & \textbf{SSIM($\uparrow$)} & \textbf{LPIPS($\downarrow$)} & \textbf{Time} \\
\midrule
Baseline (3DGSMCMC) & 22.47 & 0.830 & 0.240 & \textbf{11} & 26.11 & 0.890 & 0.140 & \textbf{22} \\
Baseline + AW-SGLD & 22.74 & 0.841 & 0.180 & 12 & 26.49 & 0.901 & 0.104 & 22 \\
Baseline + AW-SGLD + LQNAdam & \textbf{23.01} & \textbf{0.846} & \textbf{0.176} & 12 & \textbf{26.61} & \textbf{0.903} & \textbf{0.102} & 23 \\
\bottomrule
\end{tabular}
}
\caption{Ablation study on the Tanks \& Temples dataset. AW-SGLD refers to the Adaptive Weighted SGLD component and LQNAdam denotes the Local Quasi-Newton Direction-guided Adam optimizer. Time is reported in minutes.}
\label{tab:metrics}
\end{table*}

\subsection{5.2 Performance in Challenging Conditions}
We further evaluate the performance of our optimization framework Opt3DGS under various challenging conditions.

\subsubsection*{Random Initialization}

By leveraging the exploration capability of stochastic noise, 3DGSMCMC achieves good rendering quality even without using Structure-from-Motion (SfM) initialization, instead relying on random initialization. For 3DGS task, random initialization means the model starts far from high-quality solutions, thereby significantly increasing the difficulty of finding a good solution.
We report quantitative results under random initialization on all datasets in Table 2, comparing Opt3DGS with 3DGSMCMC and 3DGS. For fairness, our method uses the same random initialization scheme, training image resolution, and maximum number of Gaussians as 3DGSMCMC. Our method outperforms all baselines across all 9 metrics on the 3 datasets, showing Opt3DGS is more effective at guiding the model toward high-quality solutions even when starting from poor initial states.

\begin{figure}[ht!]
    \centering
    \includegraphics[width=\linewidth]{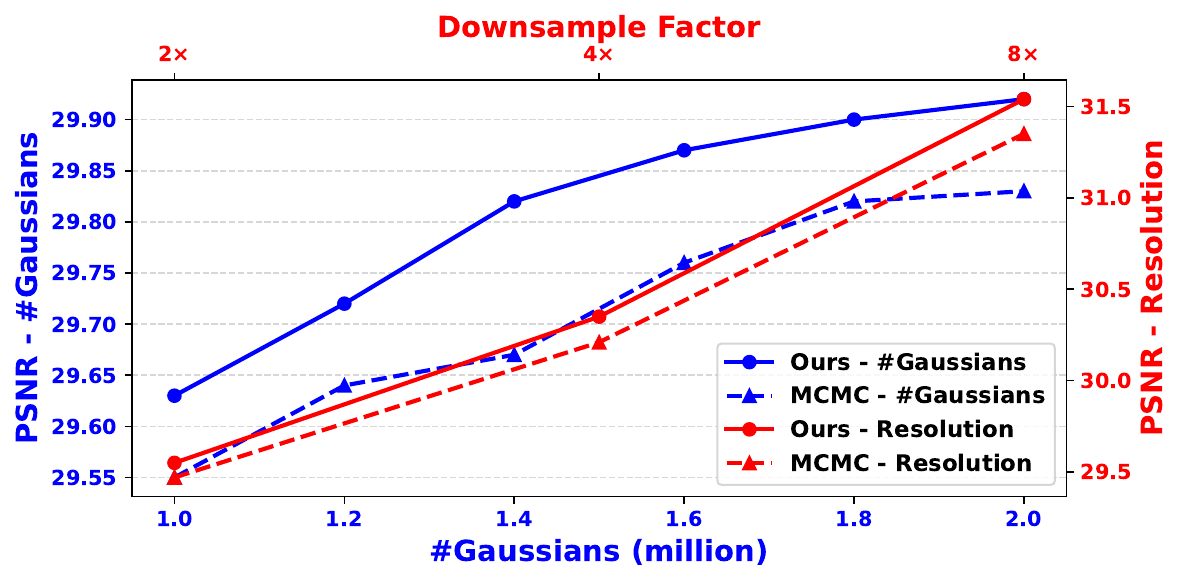}
    \caption{PSNR results on the MipNeRF dataset with different image resolutions(red) and the maximum number of Gaussians(blue).}
    \label{fig:PSNR_diffNum_and_diffRes}
\end{figure}

\subsubsection*{Higher Image Resolution}

Higher input image resolution increases the difficulty of scene fitting, as finer details and higher-frequency signals demand more accurate geometry and appearance reconstruction. This makes the posterior distribution landscape more complex and raises the risk of converging to suboptimal local modes. In Figure 4, we report PSNR comparisons between 3DGSMCMC and our method at different resolutions, on the MipNeRF360 dataset. Our method consistently outperforms the base model across all three resolution settings, demonstrating its superior robustness and effectiveness under more challenging reconstruction conditions.

\subsubsection*{Limited Number of Gaussians}

When the number of available Gaussians is reduced, the model’s representational capacity decreases. We evaluate performance under various Gaussian budget constraints on the MipNeRF360 dataset as shown in Figure 4. Our method consistently outperforms the base model, 3DGSMCMC, across all settings. This demonstrates that even with limited representational capacity, our optimization framework remains effective at guiding the model to higher-quality convergence.

\begin{figure}[ht!]
    \centering
    \includegraphics[width=\linewidth]{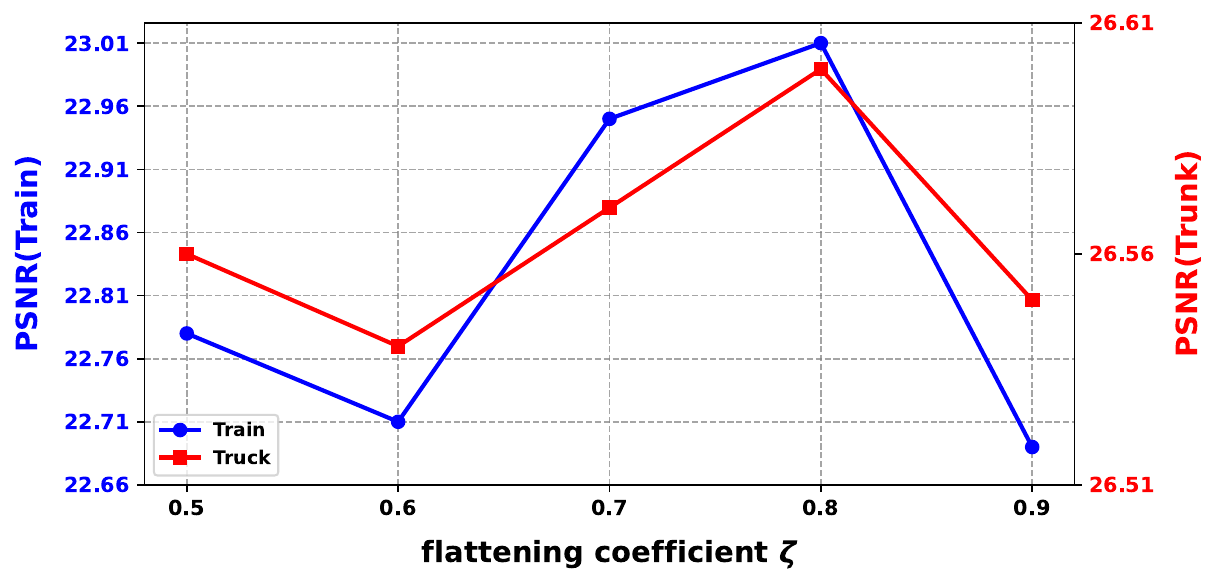}
    \caption{Ablation Study about flattening coefficient $\zeta$ on the Tanks \& Temples Dataset.}
    \label{fig:ablation_study}
\end{figure}

\subsection{5.3 Ablation Studies}

We conduct ablation experiments on the Tanks \& Temples dataset in Table 3.
All experimental settings (including image resolution and the number of Gaussians) are kept identical to those in 3DGSMCMC.
The results show that incorporating AW-SGLD alone improves rendering quality by encouraging better exploration and reducing the risk of local entrapment, while the subsequent use of LQNAdam further refines the solution and leads to higher-quality convergence through curvature-aware updates.

% \begin{table}[ht!]
% \centering
% \begin{tabular}{lcccccc}
% \toprule
% \multicolumn{1}{c}{\textbf{$\zeta$}} & \multicolumn{3}{c}{\textbf{Train}} & \multicolumn{3}{c}{\textbf{Truck}} \\
% \cmidrule(lr){2-4} \cmidrule(lr){5-7}
% \multicolumn{1}{c}{} & \textbf{PSNR} & \textbf{SSIM} & \textbf{LPIPS} & \textbf{PSNR} & \textbf{SSIM} & \textbf{LPIPS} \\
% \midrule
% 0.1 & 22.96 & 0.846 & 0.176 & 26.50 & 0.902 & 0.105 \\
% 0.2 & 22.99 & 0.847 & 0.175 & 26.51 & 0.902 & 0.103 \\
% 0.3 & 22.96 & 0.846 & 0.176 & 26.60 & 0.903 & 0.103 \\
% 0.4 & 22.87 & 0.844 & 0.178 & 26.59 & 0.903 & 0.103 \\
% 0.5 & 22.78 & 0.845 & 0.177 & 26.56 & 0.903 & 0.104 \\
% 0.6 & 22.71 & 0.843 & 0.178 & 26.54 & 0.903 & 0.104 \\
% 0.7 & 22.95 & 0.846 & 0.176 & 26.57 & 0.903 & 0.104 \\
% 0.8 & 23.01 & 0.846 & 0.176 & 26.60 & 0.903 & 0.102 \\
% 0.9 & 22.69 & 0.844 & 0.178 & 26.55 & 0.903 & 0.104 \\
% \bottomrule
% \end{tabular}
% \caption{Ablation Study about flattening coefficient $\zeta$ on the Tanks \& Temples Dataset.}
% \label{tab:metrics}
% \end{table}

 The flattening coefficient $\zeta$ is a key hyperparameter in our optimization framework, with larger values producing flatter posterior distributions. The ablation study on $\zeta$ is presented in Figure 5. We observe that Opt3DGS performs best on the Tanks \& Temples dataset when $\zeta$ values is near 0.8. This setting generalizes well across all evaluated datasets.

\section{6 Conclusion}
In this paper, we present Opt3DGS, a novel and effective optimization framework for 3DGS.
We decompose the training process into two stages—exploration and exploitation—and provide an analysis of the limitations of existing optimization strategies.
In the exploration stage, we introduce Adaptive Weighted SGLD, which enables the model to escape local minima and increases the likelihood of reaching globally optimal solutions.
In the exploitation stage, we design a Local Quasi-Newton Direction-guided Adam optimizer to achieve more accurate convergence.
Our approach improves the performance of 3DGS purely through optimization enhancements, without modifying the Gaussian representation, introducing auxiliary networks, or incurring significant additional computational costs, yet it still achieves state-of-the-art rendering quality. Looking forward, the modular nature of Opt3DGS, independent of representation or architecture, makes it a promising replacement for the optimization component in various 3DGS-based systems. Its foundation in posterior distribution reshaping and curvature-aware updates also enables the extension of optimization-centric techniques to other areas of explicit differentiable rendering.

\section*{Acknowledgements}
This research was funded by the National Natural Science Foundation of China, Grant No. 42571412.

\bibliography{aaai2026}

\end{document}